\pdfoutput=1
\documentclass[11pt]{article}

\usepackage[]{EMNLP2022}

\usepackage{times}
\usepackage{latexsym}

\usepackage[T1]{fontenc}
\usepackage[utf8]{inputenc}

\usepackage{microtype}

\usepackage{inconsolata}

\usepackage{graphicx}
\usepackage{multirow}
\usepackage{booktabs}
\usepackage{makecell}
\usepackage{amsmath}
\usepackage{enumitem}
\graphicspath{ {./pictures/} }

\title{Structure-Unified M-Tree Coding Solver for Math Word Problem}

\author{Bin Wang, Jiangzhou Ju, Yang Fan, 
Xinyu Dai\footnotemark[1], Shujian Huang, Jiajun Chen\\
National Key Laboratory for Novel Software Technology, Nanjing University \\ 
Collaborative Innovation Center of Novel Software Technology and Industrialization, Nanjing \\ 
\texttt{\{wangbin, jujiangzhou, fanyang\}@smail.nju.edu.cn} \\
\texttt{\{daixinyu, huangsj, chenjj\}@nju.edu.cn}
}

\begin{document}
\maketitle

\renewcommand{\thefootnote}{\fnsymbol{footnote}}
% 表示用9个特殊字符编号，1~9分别对应 *, +、#、§ , ¶, k, ? ?,** #。例如\footnotetext[1]{success}中的编号1对应*。
\footnotetext[1]{Corresponding author}
\renewcommand{\thefootnote}{\arabic{footnote}}  
% % 是为了把脚注标记式样恢复成标准形式（阿拉伯数字形式）。
% \footnotetext[1]{Code and data are available at }

\begin{abstract}
As one of the challenging NLP tasks, designing math word problem (MWP) solvers has attracted increasing research attention for the past few years. In previous work, models designed by taking into account the properties of the binary tree structure of mathematical expressions at the output side have achieved better performance. However, the expressions corresponding to a MWP are often diverse (e.g., $n_1+n_2 \times n_3-n_4$, $n_3\times n_2-n_4+n_1$, etc.), and so are the corresponding binary trees, which creates difficulties in model learning due to the non-deterministic output space. In this paper, we propose the Structure-Unified M-Tree Coding Solver (SUMC-Solver), which applies a tree with any M branches (M-tree) to unify the output structures. To learn the M-tree, we use a mapping to convert the M-tree into the M-tree codes, where codes store the information of the paths from tree root to leaf nodes and the information of leaf nodes themselves, and then devise a Sequence-to-Code (seq2code) model to generate the codes. Experimental results on the widely used MAWPS and Math23K datasets have demonstrated that SUMC-Solver not only outperforms several state-of-the-art models under similar experimental settings but also performs much better under low-resource conditions\footnote{Code and data are available at \url{https://github.com/devWangBin/SUMC-Solver}}. 
\end{abstract}

\section{Introduction}

\begin{figure}[ht]
\centering
\includegraphics[width=\linewidth]{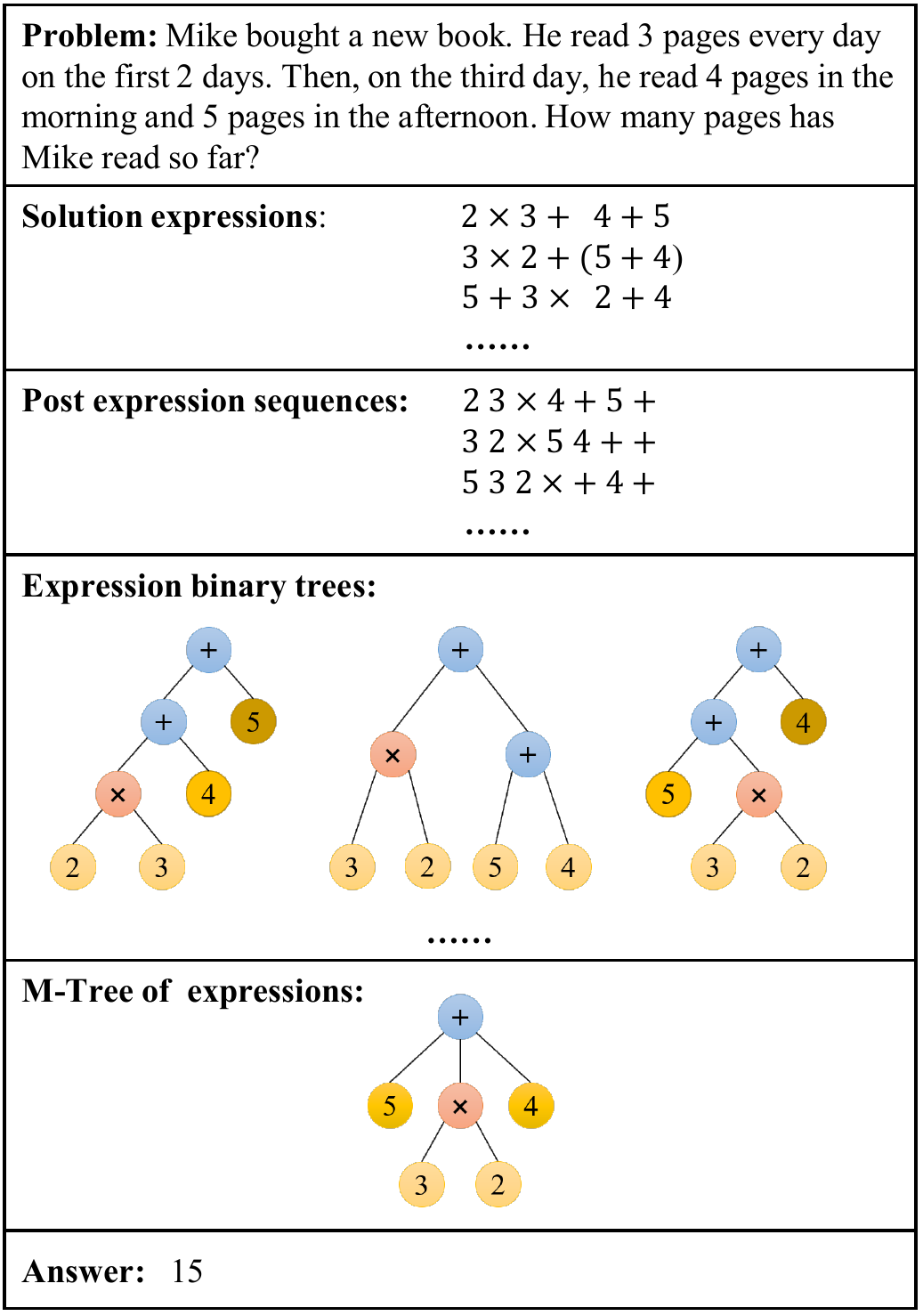}
\caption{An example of math word problems, which has multiple solution expressions and binary trees, but only one M-tree output. }
\label{fig:MWP}
\end{figure}

Given the description text of a MWP, an automatic solver needs to output an expression for solving the unknown variable asked in the problem that consists of mathematical operands (numerical values) and operation symbols ($+,-,\times , \div$), as shown in Fig. \ref{fig:MWP}. It requires that the solver not only understand the natural-language problem but also be able to model the relationships between the numerical values to perform arithmetic reasoning. 
These challenges mean MWP solvers are often broadly considered good test beds for evaluating the intelligence level of agents \citep{DBLP:conf/aaai/LinHZCLWW21}. 

% The challenges make MWP solvers be broadly considered as good test beds to evaluate the intelligence level of agents 

In recent years, the research on designing automatic MWP solvers has also made great progress due to the success of neural network models on NLP. \citet{DBLP:conf/emnlp/WangLS17} first apply a Sequence-to-Sequence (seq2seq) model to solve the MWP, and since then more methods \citep{DBLP:journals/corr/abs-1811-05632, DBLP:conf/naacl/ChiangC19, DBLP:conf/aaai/WangZZXGDS19} based on seq2seq models have been proposed for further improvement. 
To use the structural information from expressions more effectively, other methods \citep{DBLP:conf/emnlp/LiuGLK19, DBLP:conf/ijcai/XieS19} use Sequence-to-Tree (seq2tree) models with a well-designed tree-structured decoder to generate the pre-order sequence of a binary tree in a top-down manner and have achieved better performance. \citet{DBLP:conf/acl/ZhangWLBWSL20} combines the merits of the graph-based encoder and tree-based decoder (graph2tree) to generate better solution expressions. 

Promising results have been achieved in solving MWP, but the existing methods learn to output only one expression sequence or binary tree, ignoring that there is likely to be far more than one that can obtain the correct answer. 
For example, in Fig. \ref{fig:MWP}, to answer the question ``How many pages has Mike read so far?'', we can add up the number of pages that Mike read each day to get the expression ``$2\times3 +4+5$''; we can also calculate the first two days followed by the sum of pages read on the third day to get the expression ``$3\times2 + (5+4)$''; or we could even calculate the problem in a less obvious way with the expression ``$5+3\times2+4$''. 
The number of different expression sequences or binary trees can grow very large with different combinations, which results in a large non-deterministic output space and creates difficulties in model learning. 
Specifically, when a problem has multiple correct outputs, but the solver only obtains one of them, the knowledge learned by the model will be incomplete, and the demand for data will also increase, making most data-driven methods perform poorly under low-resource conditions. 

% Specifically, when each problem may have multiple legal outputs, and the solver only gets one of them to learn to solve it, the knowledge learned by the model will be incomplete, and the demand for data will also increase, which make the most data-driven methods perform poorly under low-resource conditions. 
% The challenge -- the output diversity in MWP increases the difficulty of model learning has been mentioned in previous work. 
% The output diversity problem has been mentioned in previous work. 
% The diverse output characteristic in MWP has been mentioned in previous work. 

In previous work, to overcome these limitations, \citet{DBLP:journals/corr/abs-1811-05632, DBLP:conf/aaai/WangZZXGDS19} used the equation normalization method,  which normalizes the output sequence by restricting the order of operands and has only a limited effect. 
\citet{DBLP:conf/ijcai/ZhangLLQWSS20} proposed to use multiple decoders to learn different expression sequences simultaneously. However, the large and varying number of sequences for MWPS makes the strategy less adaptable. 
For the models that learn the binary-tree output \citep{ DBLP:conf/ijcai/XieS19,DBLP:conf/emnlp/WuZFH20,DBLP:conf/acl/ZhangWLBWSL20}, they generally use a tree decoder to perform top-down and left-to-right generation that only generate one binary tree at a time, which dose not propose a solution to these limitations. 

To address the challenge that the output diversity in MWP increases the difficulty of model learning, we analyzed the causes for the diversity, which can be summarized as the following: 

\begin{itemize}[leftmargin=*]
% \begin{enumerate}
    \item Uncertainty of computation order of the mathematical operations: This is caused by 1) giving the same priority to the same or different mathematical operations. For example, in the expression $n_1+n_2+n_3-n_4$, three operations have the same priority. Consequently, the calculations in any order can obtain the correct answer, which leads to many equivalent expressions and binary trees. And 2) brackets can also lead to many equivalent outputs with different forms. For example, $n_1+n_2-n_3$, $n_1-(n_3-n_2)$ and $(n_1+n_2)-n_3$ are equivalent expressions and can be represented as different binary trees. 
    
    \item The uncertainty caused by the exchange of operands or sub-expressions: Among the four basic mathematical operations \{$+, -, \times, \div $\}, addition ``$+$'' and multiplication ``$\times$'' have the property that the operands or sub-expressions of both sides are allowed to be swapped. For example, the expression $n_1+n_2\times n_3$ can be transformed to get: $n_1+n_3\times n_2$,  $n_2\times n_3 + n_1$, etc. 
% \end{enumerate}
\end{itemize}

In this paper, to account for the aforementioned challenge, we propose SUMC-Solver for solving math word problems. The following describes the main contents of our work: 

We designed the M-tree to unify the diverse output. Existing work \citep{DBLP:conf/ijcai/XieS19,DBLP:conf/emnlp/WuZFH20,DBLP:conf/acl/WuZWH20} has demonstrated through extensive experiments that taking advantage of the tree structure information of MWP expressions can achieve better performance. 
We retain the use of a tree structure but further develop on top of the binary tree with an M-tree which contains any M branches. The ability of the M-tree to unify output structures is reflected in both horizontal and vertical directions: 
% \begin{enumerate}
\begin{itemize}[leftmargin=*]
    \item To deal with the uncertainty of computation orders for mathematical operations, we set the root to a specific operation and allow any number of branches for internal nodes in the M-tree, reducing the diversity of the tree structure in the vertical direction. 
    \item To deal with the uncertainty caused by the exchange between the left and right sibling nodes in original binary trees, we redefine the operations in the M-tree to make sure that the exchange between any sibling nodes will not affect the calculation process and treat M-trees that differ only in the left-to-right order of their sibling nodes as the same. Like the M-tree example shown in Fig. \ref{fig:MWP}. The exchange between node ``$5$'', ``$\times$'', and ``$4$'' will neither affect the calculation process nor form a new tree. With this method, the structural diversity in the horizontal direction is also reduced. 
% \end{enumerate}
\end{itemize}

We designed the M-tree codes and a seq2code framework for the M-tree learning. We abandoned the top-down and left-to-right autoregressive generation used for binary trees in previous methods. The reason is that the generation can not avoid the diversity caused by the generation order of sibling nodes. 
Instead, we encode the M-tree into M-tree codes that can be restored to the original M-tree, where the codes store the information of the paths from the root to leaf nodes and leaf nodes themselves. And inspired by the sequence labeling methods used in studies mentioned in \ref{RW-SLP}, we innovatively use a seq2code framework to generate the M-tree codes in a non-autoregressive way, which takes the problem text as the input sequence and outputs the M-tree codes of the numbers (numerical values) in the math word problem. Then we restore the codes to a M-tree that can represent the calculation logic between the numbers and finally calculate the answer. 

Our contributions can be summarized as follows: 
\begin{itemize}
    \item We analyze the causes of output diversity in MWP and design a novel M-tree-based solution to unify the output. 
    % We analyzed the causes of the output diversity of MWP and designed the M-tree to unify the output. 
    \item We design the M-tree codes to represent the M-tree and propose a seq2code model to generate the codes in a non-autoregressive fashion. To the best of our knowledge, this is the first work to analyze mathematical expressions with M-tree codes and seq2code.
    % We designed the M-tree codes to represent the M-tree equivalently and proposed a seq2code model to generate the codes in a non-autoregressive way. 
    \item Experimental results on MAWPS \citep{DBLP:conf/naacl/Koncel-Kedziorski16} and Math23K datasets \citep{DBLP:conf/emnlp/WangLS17} show that SUMC-Solver outperforms previous methods with similar settings. This is especially the case in low-resource scenarios, where our solver achieves superior performance.
    % The experimental results on the widely used Math23K dataset \citep{DBLP:conf/emnlp/WangLS17} had shown that SUMC-Solver outperforms previous methods with similar settings. Especially in low-resource scenarios, our solver can achieve much better results. 
\end{itemize}
\section{Related Work}
\subsection{Math Word Problem Solver}
\label{RW-MWP}
With the success of deep learning (DL) in various NLP tasks, designing a DL-Based MWP solver has become a major research focus lately. 
\citet{DBLP:conf/emnlp/WangLS17} first addresses the MWP with a seq2seq model, which implements the solver as a generative model from problem text sequence to expression sequence. By utilizing the semantic meanings of operation symbols, \citet{DBLP:conf/naacl/ChiangC19} apply a stack to help generate expressions. To better utilize expression structure information, other methods \citep{DBLP:conf/emnlp/LiuGLK19, DBLP:conf/ijcai/XieS19, DBLP:conf/emnlp/LiWFXXZ20} transform expressions into binary-tree-structured representations and learn the tree output. 
\citet{DBLP:conf/acl/ZhangWLBWSL20} additionally introduces a graph-based encoder to enrich the representation of problems. 

There are also approaches that explore the use of more extensive networks and external knowledge to help solve the MWP. 
\citet{DBLP:conf/acl/LiWZWDZ19} builds several special attention mechanisms to extract the critical information of the input sequence, and \citet{DBLP:conf/ijcai/ZhangLLQWSS20} propose using teacher-student networks that combine two solvers to solve the MWP. 
\citet{DBLP:conf/emnlp/WuZFH20} utilizes external knowledge through the use of an entity graph extracted from the problem sequence and \citet{DBLP:conf/aaai/LinHZCLWW21} proposes a hierarchical encoder with a dependency parsing module and hierarchical attention mechanism to make better use of the input information.
Following the previous work, \citet{DBLP:conf/acl/WuZWH20} continues to use external knowledge to enrich the problem representations and further explicitly incorporate numerical value information encoded by an external network into solving math word problems. Based on the graph2tree framework, \citet{DBLP:conf/emnlp/WuZW21} uses external knowledge to further enrich the input graph information. \citet{DBLP:conf/emnlp/YuWZX21} uses both a pre-trained knowledge encoder and a hierarchical reasoning encoder to encode the input problems.
\citet{DBLP:conf/acl/QinLHTL20} constructed different auxiliary tasks using supervised or self-supervised methods and external knowledge (common sense and problem’s part-of-speech) to help the model learn the solution of MWPs.

Different from the above methods that mainly focused on the input side and directly generated expression sequences or binary trees, we designed the structure-unified M-tree and the M-tree codes on the output side. 
Also, we design a simple model to test the advances of the M-tree and M-tree codes by comparing with methods under similar experimental settings, which means that methods using additional knowledge or multiple encoders will not become our baselines.

\subsection{Sequence Labeling Parsing}
\label{RW-SLP}
% 我们将M-tree的学习转换成M-tree codes的做法与SLP有一定的相似性，都是通过设计特殊的编码将复杂的整体结构信息转换成等价表达的编码集合。
Our method of converting the M-tree into the M-tree codes has similarities with that of sequence labeling parsing, both of which convert complex structural information into a collection of equivalently expressed codes or labels.
Constituent parsing is an NLP task where the goal is to obtain the syntactic structure of sentences expressed as a phrase structure tree. 
As the tree can be represented by a sequence of labels of the input sentence, 
\citet{DBLP:conf/emnlp/Gomez-Rodriguez18} propose transforming constituent parsing into a sequence labeling task and significantly reduce the time required for constituent parsing. 
\citet{DBLP:conf/naacl/VilaresAS19} modify the labeling scheme to avoid certain types of errors. They predict three parts of the label separately to reduce data sparsity and then combine various strategies to alleviate the error transmission problem in the original method. 
% \citet{DBLP:conf/naacl/VilaresAS19} modified the labeling scheme to avoid certain types of errors, predicted three parts of the label separately to reduce data sparsity, and combined various strategies to alleviate the error transmission problem in the original method. 
For discontinuous constituent parsing, the experiments \citep{DBLP:conf/emnlp/VilaresG20} show that despite the architectural simplicity, under the suitable representation, the sequence labeling can also be fast and accurate. \citet{DBLP:conf/naacl/StrzyzVG19} propose using a similar sequence labeling method for dependent parsing, and \citet{DBLP:conf/acl/StrzyzVG19} combine constituent parsing labeling and dependent parsing labeling with training a multi-task learning model that can perform both parsing tasks with higher accuracy. 

In the above work, the labels are used for the classification task, where the output is a one-hot vector, and each token in the input sequence corresponds to a single label. 
In contrast, our model only learns the codes of the numbers in the input sequence, where the codes are represented as non-one-hot vectors because each number may have multiple codes. Also, these codes cannot be obtained directly from the problem definition, making the design of the M-tree codes challenging.

\begin{figure*}[ht]
\centering
\includegraphics[width=\linewidth]{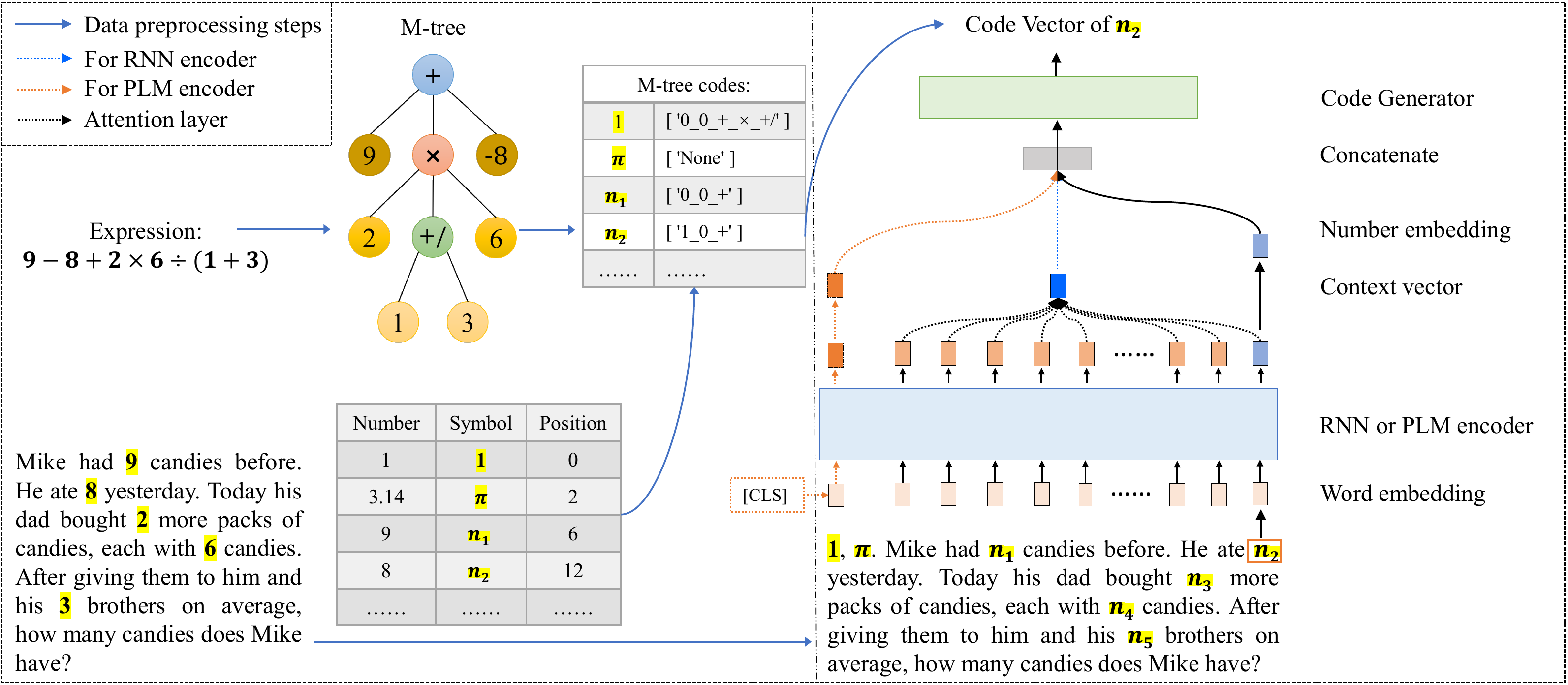}
\caption{The left half is an example of MWP with the M-tree and M-tree codes, and the right half is the main architecture of our seq2code model (see Section \ref{Design-M-tree} and Section \ref{sec-M-tree-codes-seq2code} for more details). } 
\label{fig:getMtree}
\end{figure*}
\section{The Design of SUMC-Solver}
In this section, we present the design and implementation details regarding our proposed SUMC-Solver, including the problem definition in Section \ref{PD}, the design of the M-tree in Section \ref{Design-M-tree}, and the detailed description of the M-tree codes and the seq2code model in Section \ref{sec-M-tree-codes-seq2code}. 

\subsection{Problem Definition}
\label{PD}
A math word problem is represented by a sequence of tokens, where each token can be either a word (e.g., ``Mike'' and ``candies'' in Fig. \ref{fig:getMtree}) or a numerical value (e.g., ``9'' and ``8''). Some external constants, including $1$ and $\pi$, which are required to solve the math word problem, but not mentioned in the problem text, are also added to the sequence to get the final input sequence  $X=(x_1,x_2,...,x_n)$. All the numerical values (including added constants) that appear in $X$ are denoted as a set $V=\{v_1,v_2,...,v_m\}$, and our goal is to generate a set ${C}=\{\mathbf{c}_1,\mathbf{c}_2,...,\mathbf{c}_m\}$, where $\mathbf{c}_i$ is a target code vector for $v_i$. 

\subsection{M-Tree}
\label{Design-M-tree}
\paragraph{Data Pre-processing}
For the input sequence, we add additional constants (e.g., $1$ and $\pi$) that may be used to the front of the input sequence and replace each numerical value $v_i$ with a special symbol. 
For the given expression in dataset, we try to remove all the brackets of the expression by using the SymPy\footnote{\url{https://www.sympy.org/}} Python package to prepare for the conversion of expression to the M-tree. For example, $n_1+(n_2\pm n_3)$ is converted to $n_1+n_2\pm n_3$ and $n_1\times(n_2\pm n_3)$ is converted to $n_1\times n_2 \pm n_1\times n_3$. 
For mathematical operations other than $\{+, -, \times, / \}$, such as $a^b$, we convert it to a product of multiple operands, which allows the M-tree to be extended to solve more complex mathematical problems. 

\paragraph{The Design of M-Tree}
We define the M-tree as follows: M-tree is a tree with only two kinds of nodes: internal nodes and leaf nodes, and each internal node has any M branches, where M is an integer greater than or equal to $1$. There are four types of leaf nodes, corresponding to four forms of the numerical value: $\{v, -v, \frac{1}{v}, -\frac{1}{v}\}$, which denote the original value $v$ in the problem $X$, the opposite of $v$, the reciprocal of $v$, and the opposite of the reciprocal of $v$, respectively. There are four types of internal nodes, corresponding to four redefined operations $\{+, \times, \times -, +/ \}$  that ensure sibling nodes are structurally equivalent in the M-tree and two M-trees that differ only in the order of their sibling nodes will be treated as the same. The root of the M-tree is set as a ``$+$'' node to unify the structure (operators can have only $1$ operand, so $n_1\times n_2$ will be represented as a unique subtree of the root node). 
For an internal node that has $k$ children $\{v_1,v_2,...,v_k\}$, where $k$ is an integer greater than or equal to $1$: 
\begin{itemize}
    \item The node of ``$+$'' (``$\times $'') means to sum (multiply) the values of all its child nodes: $v_1+v_2+,...,+v_k$ \  ($v_1\times v_2\times ,...,\times v_k$). 
    \item The node of ``$\times -$''(``$+/$'') means to get the opposite (reciprocal) of the product (sum) value of all its child nodes: $-v_1\times v_2\times ,...,\times v_k$ \  ($\frac{1}{v_1+v_2+,...,+v_k}$). 
\end{itemize}
The implementation details of the M-tree are provided in Section \ref{sec:appendix} in the Appendix. 

\subsection{M-Tree Codes and Seq2code Model}
\label{sec-M-tree-codes-seq2code}

\subsubsection{The Design of M-Tree Codes}
\label{M-Tree Codes}
Since the nodes in the M-tree can have any number of branches and sibling nodes are structurally equivalent, autoregressive-based generation cannot avoid the diversity caused by the sequential order of sibling nodes at the output side.
To address this challenge, we encode the structure information of the M-tree into each leaf node, forming a mapping between the M-tree and the codes set of leaf nodes so that the model can generate the codes in a non-autoregressive way. Details about M-tree codes are as follows: 

\paragraph{Components of M-tree Codes}
The M-tree code of each leaf node consists of two parts: one part describes the numerical value, and the other part is formed by the path from the root node to the current leaf node. A specific example is shown in Fig. \ref{fig:getMtree}. The first part of the code uses two binary bits to distinguish the four forms (mentioned in \ref{Design-M-tree}) of numerical values. Specifically, for a leaf node in the M-tree represented as $v_i^{'}$, where $v_i$ is the numerical value in the input sequence, the first part of the M-tree code of $v_i^{'}$ will be set according to the following rules: 
 
{ • } If $v_i^{'} = v_i$,  the code is set as ``$0\_0$'';

{ • } If $v_i^{'} = -v_i$,  the code is set as ``$1\_0$'';

{ • } If $v_i^{'} = \frac{1}{v_i}$,  the code is set as ``$0\_1$'';

{ • } If $v_i^{'} = -\frac{1}{v_i}$,  the code is set as ``$1\_1$'';\\
The second part is set as the sequential operation symbols of all internal nodes on the path from the root to the current leaf node $v_i^{'}$, so leaf nodes with the same parent node will share the same second part code. 
For example, the second part of the M-tree code of ``$-8$'' in the example showing in Fig. \ref{fig:getMtree} is ``$+$'', and the code of "$1$" or ``$3$'' is ``$+\_\times \_+/$''. 
In some special cases, if the internal nodes that are siblings have the same type (e.g., all ``$\times$'' nodes), they need to be marked with a special symbol added to the end to distinguish them from each other in order to restore the correct M-tree from the codes. 

After converting all M-trees in the training dataset to M-tree codes, a set of M-tree codes will be obtained.
% and in order to increase generalization, we can add new ones to expand the existing M-tree codes. 
The final set of M-tree codes is denoted as $B=\{b_1,b_2,...,b_l\}$, which has $l$ different codes in total. 
For example, in the example of Fig. \ref{fig:getMtree}, the M-tree code ``$1\_0\_+$'' of ``$-8$'' is an element of $B$. 

\paragraph{Vector Representation of M-tree Codes}
The final code vector $c_i$ for model learning will be obtained based on $B$. 
Considering that the value $v_i$ that appears only once in the input problem text may appear multiple times in the M-tree. For example, in ``$v_i\times v_j \pm v_i \times v_k$'', $v_i$ will appear in two leaf nodes and have two identical or different M-tree codes. 
Consequently, the set of numerical values $V=\{v_1,v_2,...,v_m\}$ is mapping to a set of  $l$-dimensional non-one-hot vectors: $\mathbf{C}=\{\mathbf{c_1,c_2,...,c_m}\}$, where $\mathbf{c_i}$ is the code vector of the corresponding ${v_i}$ and the value of $\mathbf{c_i}$ in the $k$-th dimension indicates how many codes of $b_k$ that $v_i$ has.  For example, the final code vector of the value ``$\pi$'' in the example showing in Fig. \ref{fig:getMtree} will be set as $\left[1,0,...,0\right]^{\top}$, where only the first dimension has the value of $1$ indicating that ``$\pi$'' has only one M-tree code ``None'', which means that it does not appear in the M-tree. 

\paragraph{Reducing M-tree codes to M-tree}
The process of converting M-tree to M-tree codes is reversible. Briefly, a code vector is generated for each number in the text and mapped to one or more M-tree codes at first. Then, the number is formatted according to the first part of the M-tree code. Finally, all the numbers are merged by scanning the second part of the M-tree code from back to front, while the M-tree is generated bottom-up. 

\subsubsection{Sequence-to-Code Model}
To verify the advances of the M-tree and M-tree codes, we design a simple seq2code model to tackle the MWP task, which takes the problem sequence as its input and then outputs the corresponding codes (represented as vectors) for numerical values in the problem. After combining all the codes to restore the M-tree, we can calculate the final answer for the problem. Next, we introduce the two core parts of the model: the problem encoder and the code generator. 

\paragraph{Problem Encoder}
We use an encoder to transform the words of a MWP into vector representations. There are two kinds of encoders used in our experiments: a Recurrent Neural Network (RNN) encoder or a pre-trained language model (PLM) encoder. 

For the RNN encoder, we use a bidirectional LSTM (BiLSTM) \cite{DBLP:journals/neco/HochreiterS97} network. Formally, given the input sequence $X=(x_1,x_2,...,x_n)$ and the numerical values set $V=\{v_1,v_2,...,v_m\}$, we denote the positions of the numerical values as $Q=\{q_1,q_2,...,q_m\}$, in which $q_i$ is the position of $v_i$ in $X$.
The encoder encodes the input sequence into a sequence of hidden states $\mathbf{H}=\{\mathbf{h}_1^{x},\mathbf{h}_2^{x},...,\mathbf{h}_n^{x}\} \in \mathrm{R}^{n \times 2 d} $ as follows:

\begin{equation} 
\label{BiLSTM-encoder}
\begin{array}{ll}
\mathbf{h}_{t}^{x}=\left[\overrightarrow{\mathbf{h}_{t}^{x}}, \overleftarrow{\mathbf{h}_{t}^{x}}\right], \vspace{1ex} \\ 
\overrightarrow{\mathbf{h}_{t}^{x}},\overrightarrow{\mathbf{c}_{t}^{x}}={BiLSTM}\left(\mathbf{e}_{t}^{x},\overrightarrow{\mathbf{c}_{t-1}^{x}}, \overrightarrow{\mathbf{h}_{t-1}^{x}}\right), \vspace{1ex} \\
\overleftarrow{\mathbf{h}_{t}^{x}}, \overleftarrow{\mathbf{c}_{t}^{x}}={BiLSTM}\left(\mathbf{e}_{t}^{x},\overleftarrow{\mathbf{c}_{t-1}^{x}}, \overleftarrow{\mathbf{h}_{t-1}^{x}}\right).
\end{array}
\end{equation}
Where 
$\mathbf{e}_{t}^{x}$ is the word embedding vector for $x_t$, $n$ is the size of input sequence $X$, $d$ is the size of the LSTM hidden state, and $\mathbf{h}_{t}^{x}$ is the concatenation of the forward and backward hidden states. 

And then for the numerical value $v_i$ in the problem $X$, its semantic representation $\mathbf{e}_i^{c}$ is modeled by the corresponding BiLSTM output vector: 
\begin{equation}
    \mathbf{e}_{i}^{c} = \mathbf{h}_{q_i}^{x} .
\end{equation}
In order to better capture the relationship between different numerical values and the relationship between $v_i$ and the unknown value to be solved (answer of the problem), we use an attention layer to derive a context vector $\mathbf{E}_{i}$ for $v_i$, which is expected to summarize the key information of the input problem and help generate the final target code for $v_i$. The context vector $\mathbf{E}_{i}$ is calculated as a weighted representation of the source tokens:

\begin{equation}
    \mathbf{E}_{i}=\sum_{t} \alpha_{i t} \mathbf{h}_{t}^{x}, 
\end{equation}
where 
\begin{center}
\begin{equation*}
    \alpha_{i t}=\frac{\exp \left(\operatorname{score}\left(\mathbf{e}_{i}^{c}, \mathbf{h}_{t}^{x}\right)\right)}{\sum_{t} \exp \left(\operatorname{score}\left(\mathbf{e}_{i}^{c}, \mathbf{h}_{t}^{x}\right)\right)}
\end{equation*}
\end{center}
and 
\begin{center}
\begin{equation*}
    \operatorname{score}\left(\mathbf{e}_{i}^{c}, \mathbf{h}_{t}^{x}\right)=\mathbf{U}^{\top} \tanh \left(\mathbf{W}\left[\mathbf{e}_{i}^{c}, \mathbf{h}_{t}^{x}\right]\right).
\end{equation*}
\end{center}
where $\mathbf{U}$ and $\mathbf{W}$ are trainable parameters. Finally, we concatenate context vector $\mathbf{E}_{i}$ and $\mathbf{e}_i^{c}$ to obtain $\mathbf{z}_{i}^{c}$ as the input of the generator: 
\begin{equation}
    \mathbf{z}_{i}^{c} = \left[\mathbf{E}_{i}, \mathbf{e}_i^{c}\right].
\end{equation}

For the PLM encoder, we use RoBERTa-base \citep{DBLP:journals/corr/abs-1907-11692} or BERT-base \citep{DBLP:conf/naacl/DevlinCLT19} to encode the input sequence $X$ to get the token embeddings $Ems =\left\{em_{t}^{x}\right\}_{t=1}^{n}$ and get the semantic representation $\mathbf{e}_{i}^{c}$ in the same way as the RNN encoder, but for the context vector $\mathbf{e}_{i}^{c}$ we use the output embedding of the special token [CLS] in RoBERTa. 
\begin{align}
    \mathbf{e}_{i}^{c} & = \mathbf{em}_{q_i}^{x}, \\
    \mathbf{E}_{i} & = \mathbf{em}_{cls}^{x} .
\end{align}

\paragraph{Code Generator}
We use a simple three-layer Feedforward Neural Network (FFNN) to implement the generator. With the input $\mathbf{z}_{i}^{c}$, the final code vector $\mathbf{c}_{i}^{'}$ is generated as follows: 
\begin{equation}
\begin{split}
\mathbf{z}_{i 1}^{c}& =  \mathbf{\sigma}\left({\mathbf{z}_{i}^{c}}^{\top} \mathbf{W}_{1} + \mathbf{B}_{1}  \right), \vspace{1ex} \\
\mathbf{z}_{i 2}^{c} & =  \mathbf{\sigma}\left({\mathbf{z}_{i 1}^{c}}^{\top} \mathbf{W}_{2} + \mathbf{B}_{2}  \right), \vspace{1ex} \\
\mathbf{c}_{i}^{'}& =  {\mathbf{z}_{i 2}^{c}} ^{\top} \mathbf{W}_{3} + \mathbf{B}_{3}.
\end{split}
\end{equation}
Where $\mathbf{\sigma}$ is an activation function, $\mathbf{W}_{i}$ and $\mathbf{B}_{i}$ are the 
parameters of the FFNN. 
% and $\mathbf{c}_{i}^{'}$ is the generated code vector. 

\paragraph{Training Objective}
Given the training dataset $\mathbf{D}=\{\left(X^{i}, C^{i}\right): 1 \leq i \leq N \}$, 
where $C^{i}$ is the set of all the code vectors corresponding to the numerical values appearing in $X^{i}$, we minimize the following loss function:
\begin{equation}
    \mathcal{L}= \sum_{(X^{i}, C^{i}) \in \mathbf{D}} \sum_{\mathbf{c}_{i} \in C^{i}} \mathcal{L}_{MSE} (\mathbf{c}_{i}, \mathbf{c}_{i}^{'}), 
\end{equation}
where  
\begin{equation}
    \mathcal{L}_{MSE} (\mathbf{c}_{i}, \mathbf{c}_{i}^{'}) = \frac{1}{l} \sum_{j=1}^{l}\left(\mathbf{c}_{i j}-\mathbf{c}_{i j}^{'}\right)^{2}. 
\end{equation}
where $l$ is the dimensionality of code vectors. 
\section{Experiments}
\subsection{Datasets}
We evaluate our SUMC-Solver on two commonly used MWP datasets, MAWPS \citep{DBLP:conf/naacl/Koncel-Kedziorski16} with 2,373 problems and Math23K\footnote{Available from \url{https://ai.tencent.com/ailab/nlp/dialogue/##Dataset/}} with 23,162 problems.
For Math23K, we use the public test set. For MAWPS, we evaluate the performance via five-fold cross-validation and improved the pre-processing method in the previous work \citep{DBLP:conf/ijcai/XieS19, DBLP:conf/acl/ZhangWLBWSL20} to avoid coarsely filtering out too much data, and the final amount of available data was 2,373 (previously 1,921).
We use answer accuracy as the evaluation metric: if the value predicted by the solver equals the true answer, it is thought of as correct. 

% 可能不需要
\subsection{Implementation Details}
The parameter settings are as follows: 
1) For the RNN encoder, the dimensionality of word embedding and hidden states are $128$ and $512$, respectively. We select nearly 2500 words that appear most frequently in the training set as the vocabulary and replace the remaining words with a unique token UNK. The global learning rates are initialized to $0.002$ for Math23K and $0.008$ for MAWPS. 2) For the PLM encoder, we use RoBERTa-base and BERT-base for Math23K and MAWPS, respectively. The initial global learning rate for both datasets is $2\times 10^{-5}$. 
3) For the code generator, the dimension of the FFNN is (2048, 1024, $|c_i|$), where $c_i$ is the code vector and its dimensionality is $153$ for Math23K and $28$ for MAWPS, respectively.

\subsection{Compared Methods}
Considering SUMC-Solver with one traditional sequence encoder without any other external knowledge as input and one simple generator, we only compare methods with similar settings: 
\textbf{T-RNN} \citet{DBLP:conf/aaai/WangZZXGDS19} applied a seq2seq model to predict a tree-structure template, which includes inferred numbers and unknown operators. Then, They used a RNN to obtain unknown operator nodes in a bottom-up manner. 
\textbf{StackDecoder} \citet{DBLP:conf/naacl/ChiangC19} used the RNN to understand the semantics of problems, and a stack was applied to generate post expressions. 
\textbf{GTS} \citet{DBLP:conf/ijcai/XieS19} utilized a RNN to encode the input and another RNN to generate the expression based on top-down decomposition and bottom-up subtree embedding. 
\textbf{GTS-PLM} replaces the encoder with a pre-trained language model compared to the original GTS. 
\textbf{SAU-Solver} \citet{DBLP:conf/emnlp/QinLLZL20} devised Universal Expression Trees to handle MWPs with multiple unknowns and equations. Then a RNN encodes the input and a well-designed decoder considering the semantic transformation between equations obtains the expression. 
\textbf{Graph2Tree} \citep{DBLP:conf/acl/ZhangWLBWSL20} is a graph-to-tree model that leverages an external graph-based encoder to enrich the quantity representations in the problem. 
\textbf{UniLM-Solver} UNIfied Pre-trained Language Model~(UniLM)~\citep{DBLP:conf/nips/00040WWLWGZH19} have achieved superior performance on natural language understanding and generation tasks, which can be used to model the generation process from the input text to the output expression. 

\subsection{Results and Analyses}
\paragraph{Answer Accuracy} 
The experiment results are shown in Table~\ref{tab:main-acc}. We observe that SUMC-Solver outperforms all baselines in the two MWP datasets. 
When using an RNN as the encoder, SUMC-Solver surpasses StackDecoder and T-RNN that learn the sequence output by 9-10 percent. For methods that learn the binary-tree output, SUMC-Solver also achieves better results than GTS, SAU-Solver and Graph2Tree, although these methods used a well-designed tree decoder or an external graph-based encoder to enrich the representations. 
When using a PLM as the encoder, SUMC-Solver achieves an accuracy of $82.5\%$, a significant improvement ($3$ and $5$ percent, respectively) over GTS-PLM and UniLM-Solver. 
In conclusion, the two different encoder settings above both show that the design of the M-tree and M-tree codes is reasonable and advanced, which allows us to achieve better performance using only a simple seq2code model. 

\begin{figure}[tbp]
\centering
\includegraphics[width=\linewidth]{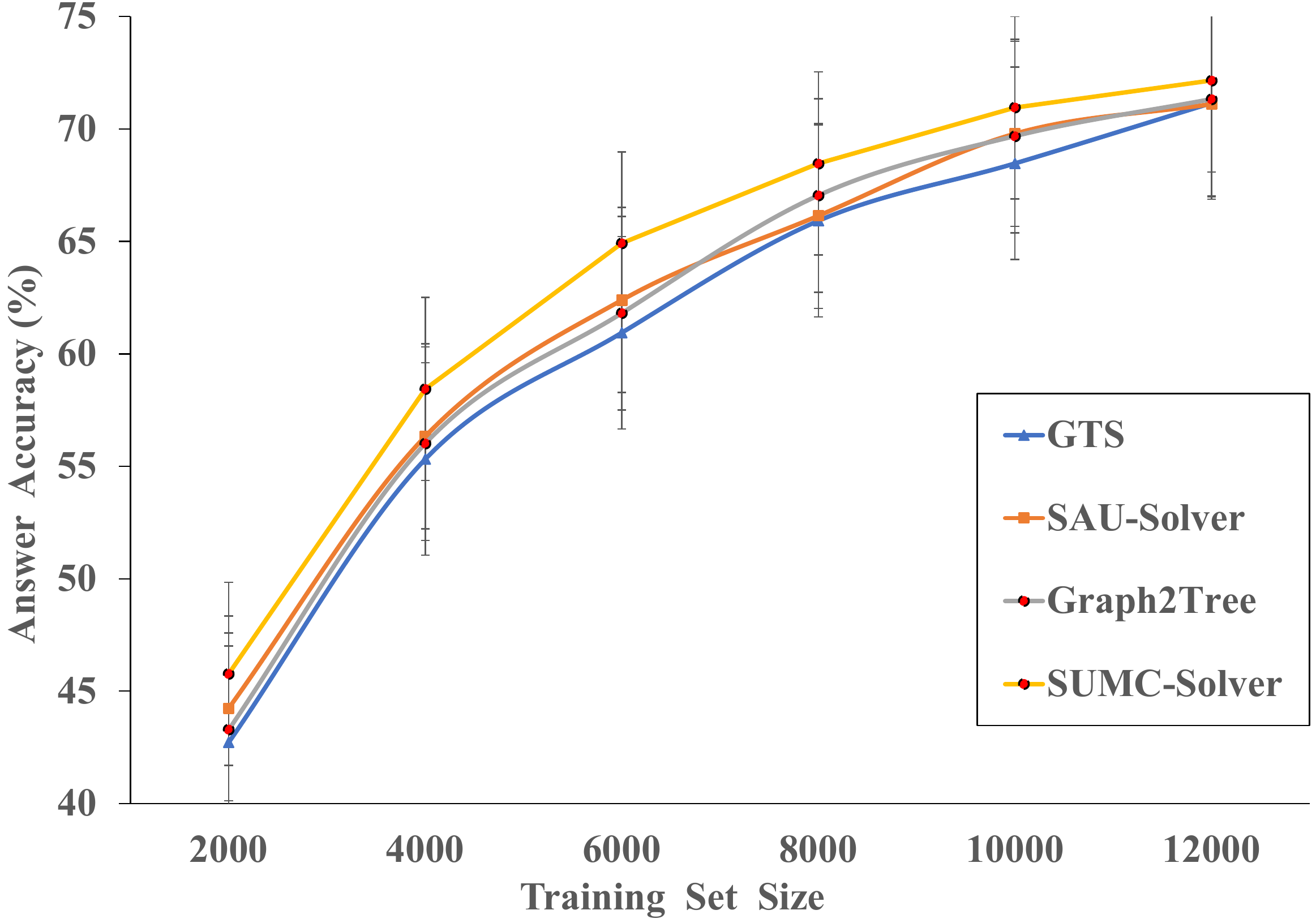}
\caption{Answer accuracy in different low-resource conditions}
\label{fig:low-acc}
\end{figure}

\paragraph{Comparison in Low-resource Situations}
% in low-resource scenarios
% 标注成本是昂贵的, 模型在低资源下的表现也是重要的.
% a. 在所有的训练集大小设置下, 都取得最好的性能. 
% b. 25%, 可以取得, 在低资源下有更大性能的优势.比GTS高5个点.
% c. 比参数更少的GTS学习的更快.
% so we hope the model performs well in lower resources.
The annotation cost for MWPs is high, so it is desirable for the model to perform well in lower resource settings. 
Therefore, we evaluate our model performance with GTS, SAU-Solver and Graph2Tree on training sets of different sizes. 
The test set contains 2,312 randomly sampled instances. Detailed results can be found in Fig. \ref{fig:low-acc}. 
Tt can be observed that SUMC-Solver consistently outperforms other models irrespective of the size of the training set. 
Firstly, when the size of the training set is less than $6000$, the performance of SAU-Slover is better than that of GTS; when the number exceeds $6000$, these two models perform similarly. In terms of overall performance, the results of SAU-Solver and Graph2Tree are better than those of the GTS when resources are constrained. 
Secondly, with a 6000-sample training set, the most significant performance gap between SUMC-Solver and other models occurs, where our model approximately obtains an additional $5\%$ on accuracy. This shows that SUMC-Solver has the most prominent advantages in low-resource situations. 

\begin{figure}[tbp]
\centering
\includegraphics[width=\linewidth]{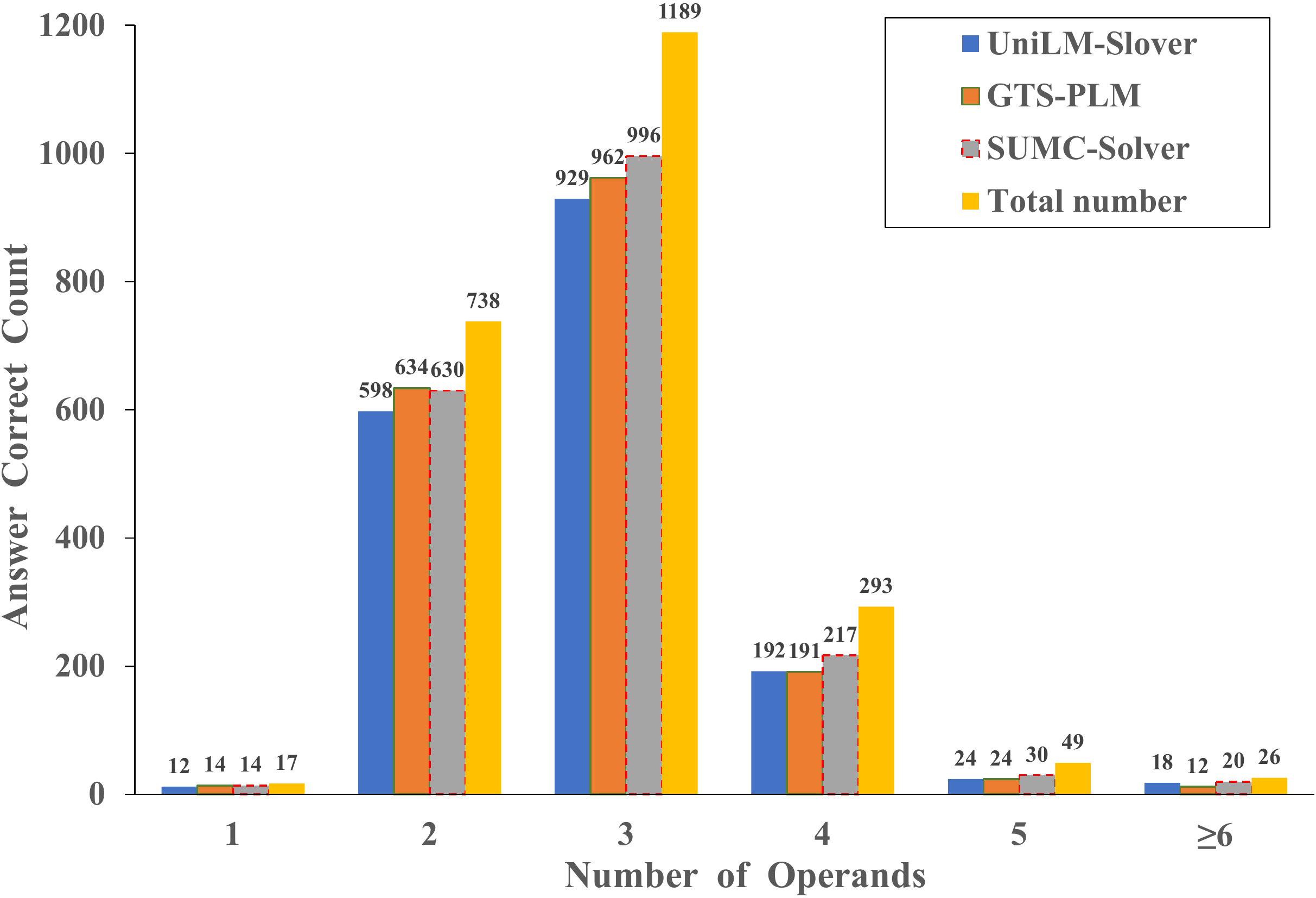}
% 模型的性能在要求不同操作数数量的数据上的比较
\caption{Comparison of the number of problems answered correctly by different models on test data, where the test data are classified according to the number of operands they require. }
\label{fig:num-numbers}
\end{figure}

\begin{table}[tbp]
\centering
\begin{tabular}{l|l|l|l} 
\toprule
\multicolumn{1}{l|}{}& Model        & Math23K       &MAWPS* \\ \midrule
\multirow{4}{*}{RNN}                                                                                     
                    & T-RNN         & 66.9          &66.8   \\
                    & StackDecoder  & 67.8          &-      \\
                    & GTS           & 75.6          &75.2$^{\dagger}$   \\
                    & SAU-Solver    & 76.2$^{\dagger}$          &75.5$^{\dagger}$   \\
                    & Graph2Tree    & 76.6$^{\dagger}$          &78.1$^{\dagger}$   \\
                    & SUMC-Solver      & \textbf{77.4} &\textbf{79.9}   
                                \\ \midrule
\multicolumn{1}{l|}
{\multirow{3}{*}{PLM}} 
                    & UniLM-Solver  & 77.5$^{\dagger}$     &78.0$^{\dagger}$\\
\multicolumn{1}{l|}{}                                                     
                    & GTS-PLM &79.5$^{\dagger}$      &79.8$^{\dagger}$\\
\multicolumn{1}{l|}{}                                
                    & SUMC-Solver & \textbf{82.5}      &\textbf{82.0}\\ 

\bottomrule

\end{tabular}
% \caption{The comparison of math word problem solvers on Math23K dataset}
\caption{
% 是否要说明 那些是复现的结果 % Model comparison on answer accuracy via 5-fold cross-validation.
Answer accuracy of SUMC-Solver and various baselines. Math23K denotes results on the public test set, MAWPS* denotes 5-fold cross-validation and the results with $^{\dagger}$ are obtained by our reproduction. We reproduced the results for: 1) Getting new results, such as the results of SAU-Solver on the public test set of Math23K and the results of GTS-PLM; 2) Using improved data preprocessing method for MAWPS. }
\label{tab:main-acc}
\end{table}

% \paragraph{Impact of Number of Operands}
\paragraph{Performance on Different Numbers of Operands}
We divide the test set (2,312 randomly sampled instances) into different levels according to the number of operands (numerical values in problems) required to calculate the answer and evaluate the model performance on these different data. The details are shown in Fig. \ref{fig:num-numbers}. 
From the results, we can see that most of the MWPs require between $2$ and $4$ operands, and SUMC-Slover performs better than the baseline models on data requiring more operands, which shows that our solver has the potential to solve more complex problems. 

\begin{table}[tbp]
\centering
\begin{tabular}{ccc} 
\toprule
\multicolumn{1}{c}{Codes} 
& \makecell[c]{Code set size}             
& \makecell[c]{Test set \\coverage (\%)} \\ 
\midrule
\multirow{1}{*}{M-tree}                                                                                                       & 153        & 100.0    \\
\multirow{1}{*}{Binary-tree}                                                                                                  & 1290        & 93.5   \\
\bottomrule
\end{tabular}
\caption{Comparison of Binary-Tree and M-Tree Codes. }
\label{tab:acc-code}
\end{table}

% \begin{table}[ht]
% \centering
% \begin{tabular}{|c|c|c|}
% \hline
% Encoder                & \makecell[l]{Accuracy of the\\ code vectors (\%)} & \makecell[l]{Answer Accuracy \\of MWPs (\%)} \\ \hline
% RNN & 88.4                         & 74.9             \\ \hline
% PLM & 92.7                         & 82.5             \\ \hline
% \end{tabular}
% \caption{Accuracy of the code vectors.}
% \label{tab:acc-code}
% \end{table}

\paragraph{Comparison of Binary-Tree and M-Tree Codes}
% 从结果可以观察到，应用M-tree的结构能极大地减少编码的大小并且能够保证所得到的编码集能够覆盖测试集中的数据，这表明M-tree对于输出结构进行统一的效果是非常显著的。
The seq2code framework can also be applied to the binary-tree structure if choosing one binary tree for each MWP and converting it to the codes in the same way. We transformed the data of Math23K's training set and compared the binary-tree codes and M-tree codes, which is shown in the Table \ref{tab:acc-code}. 
It can be observed that applying the M-tree structure can greatly reduce the size of the code set and ensure that the obtained codes can cover the data in the test set, which shows the effect of the M-tree on unifying the output structure is very significant. 

% From Table \ref{tab:acc-code}, we observe that our solver can achieve an accuracy of about $90\%$ in the code vectors learning, but it is difficult to achieve such high accuracy of the answer obtained by combining all the code vectors. The reason is that each problem contains multiple numerical values, and the model must output each corresponding code vector correctly to get the correct answer. At present, the code vector is obtained by rounding the output and the restoration from the code vectors to the M-tree is based on simple rules, and more feasible methods can be explored in the future. 

\section{Conclusion}
In this paper, we proposed SUMC-Slover to solve math word problems, which applies the M-tree to unify the diverse output and the seq2code model to learn the M-tree. 
The experimental results on the widely used MAWPS and Math23K datasets demonstrated that SUMC-Solver outperforms several state-of-the-art models under similar settings and performs much better under low-resource conditions. 

\section*{Limitations}
Some discussions on the limitations of SUMC-Solver are as follows: 
1) The M-tree corresponding to the output of each MWP is unique. However, as mentioned in Section \ref{M-Tree Codes}, some special M-trees need to be distinguished by introducing special symbols randomly when converting them into M-tree codes, which makes the M-tree codes correspond to the MWP may not be unique. 
Through the statistics of the datasets, we found that about $90\%$ of the data do not belong to this particular case. At the same time, for the remaining $10\%$, despite the increased difficulty, they are still learnable based on previous work experience, which makes SUMC-Solver still achieve a significant performance improvement. 
2) The network structure is relatively simple for the seq2code framework used in SUMC-Solver. 
In previous work, the use of graph-based encoders and the introduction of external knowledge to enrich the representation of the input problem text have been shown to further improve the performance of the solver, and seq2code can be naturally integrated with these improved approaches to try to achieve better results. 

\section*{Acknowledgements}
We would like to thank the anonymous reviewers for their constructive comments. This work was supported by the National Natural Science Foundation of China (No. 61936012 and 61976114).

\bibliography{custom}
\bibliographystyle{acl_natbib}

\appendix

\section{Implementation Details of the M-tree}
\label{sec:appendix}
% 所得到的表达式中的每个运算符所关联的操作数数量将会被降到最低2个，除了两种特殊情况：
After the data pre-processing for expression mentioned in \ref{Design-M-tree}, We can easily convert it into a M-tree based on the following steps: 

% \begin{enumerate}
(1) By following the order of priority for operations: (operations in brackets) $> (\times=\div) > (+=-)$, Converting the operations one-by-one in the expression as follows:
% ``$-$'' and ``$\div$'' in expressions. 
% Handling the special case $v_1\div(v_2 \pm v_3)$, which is converted to $n_1\times n_2  $
\begin{enumerate}
    \item For $v_1 \div v_2$, it is converted to $v_1 \times v_2^{'}$, where $ v_2^{'} $ is the reciprocal of $v_2$. 
    \item For $v_1 - v_2$, it is converted to $v_1 + v_2^{'}$, where $v_2^{'}$ is the opposite of $v_2$. 
    \item For $v_1 - v_2\times v_3$, it is converted to $v_1 + v_2 (\times -) v_3$, where $v_2 (\times -) v_3$ means the opposite of $v_2\times v_3$. 
    \item For $v_1 \div( v_2 + v_3)$, it is converted to $v_1 \times v_2 (+ /) v_3$, where $v_2 (+ /) v_3$ means the reciprocal of $v_2 + v_3$. 
\end{enumerate}
% After the above conversion is completed, a new expression is obtained, which only contains the four operations defined in M-tree ().  
After the conversion, only four operations we defined in the M-tree will be left in the new expression, and they all have the property that the computation is not affected by the left-right order between child nodes, which can be used to reduce the structural diversity in the horizontal direction. 
% In addition, in this paper, two M-trees that differ only in the order of their sibling nodes will be defined as the same. 
% 进行分析，对于新得到的二叉树...优势，优点，对比以前的二叉树
% 对于上述操作的第一个问题是：为什么要抛弃 - 和 / ？ 因为想去确保左右兄弟节点对于父节点的运算是等价的操作数，一来是：无论如何调换兄弟节点的位置，所执行的都是同一种计算，提高M-tree的结构泛化性; 二来是：在对于叶子节点进行编码时，可以给兄弟节点赋予相同的code，简化编码

(2) After obtaining the new expression, we convert it to a binary tree and then reduce it from top to bottom to get the final M-tree.  Let the parent node be $v_p$ and the child node be $v_c$, and the details are as follows: 
\begin{enumerate}

    \item If it is one of the $4$ cases: 
    \label{case001}
    1) ``$v_p = v_c = +$'', 2) ``$v_p=v_c= \times$'', 3) ``$v_p=+/$ and $v_c=+$'', 4) ``$v_p=\times -$ and $v_c=\times$'',
    then merge directly, delete the child node $v_c$ and assign its children (if has) to $v_p$ and continue checking down. 
    \item If ``$v_p=\times $ and $v_c=\times -$'', then make ``$v_p=\times -$'' and do the same as \ref{case001}.
    % delete $v_c$ and assign its children (if has) to $v_p$ and continue checking down.
    \item If ``$v_p=\times -$ and $v_c=\times -$'', then make ``$v_p=\times$'' and do the same as \ref{case001}.
    % delete $v_c$ and assign its children (if has) to $v_p$ and continue checking down.

\end{enumerate}
% 进行这样的合并后所得到的M-tree有什么用，优势？结合结构统一...
% After top-down merging of nodes, the height of the tree will be minimized and the tree structure will be unified vertically
After merging the nodes from top to bottom, the height of the tree will be minimized, and the tree structure will be unified in the vertical direction. 
And we obtain a structure-unified M-Tree for the origin solution expression. 

% This is a section in the appendix.

\end{document}